\PassOptionsToPackage{sort&compress}{natbib}
\documentclass[a4paper, fleqn]{cas-sc}
\usepackage[numbers]{natbib}
\usepackage{graphicx}
\usepackage{booktabs}
\usepackage{mathrsfs}
\usepackage[ruled, vlined]{algorithm2e}
\usepackage{array}
\usepackage[caption=false, font=normalsize, labelfont=sf, textfont=sf]{subfig}
\usepackage{pifont}
\usepackage{tabularx}
\usepackage{times}
\usepackage{soul}
\usepackage{url}
\usepackage[small]{caption}
\usepackage[switch]{lineno}
\usepackage{multirow} 
\usepackage{textcomp}
\usepackage{picinpar}
\usepackage{amsthm, amsmath, amssymb}
\usepackage{microtype}
\def\tsc#1{\csdef{#1}{\textsc{\lowercase{#1}}\xspace}}
\tsc{WGM}
\tsc{QE}

\begin{document}
\let\WriteBookmarks\relax
\def\floatpagepagefraction{1}
\def\textpagefraction{. 001}

\title{Confidence-driven Gradient Modulation for Multimodal Human Activity Recognition: A Dynamic Contrastive Dual-Path Learning Approach}
\author{%
  \texorpdfstring{Panpan Ji\textsuperscript{\#}}{Panpan Ji\#}, 
  \texorpdfstring{Junni Song\textsuperscript{\#}}{Junni Song\#}, 
  Yifan Lu, Hang Xiao, Hanyu Liu, 
  \texorpdfstring{Chao Li\textsuperscript{*}}{Chao Li*}%
}
\begin{abstract}
Sensor-based Human Activity Recognition (HAR) is a core technology that enables intelligent systems to perceive and interact with their environment. However, multimodal HAR systems still encounter key challenges, such as difficulties in cross-modal feature alignment and imbalanced modality contributions.
To address these issues, we propose a novel framework called the Dynamic Contrastive Dual-Path Network (DCDP-HAR). The framework comprises three key components. First, a dual-path feature extraction architecture is employed, where ResNet and DenseNet branches collaboratively process multimodal sensor data. Second, a multi-stage contrastive learning mechanism is introduced to achieve progressive alignment from local perception to semantic abstraction. Third, we present a confidence-driven gradient modulation strategy that dynamically monitors and adjusts the learning intensity of each modality branch during backpropagation, effectively alleviating modality competition. In addition, a momentum-based gradient accumulation strategy is adopted to enhance training stability.
We conduct ablation studies to validate the effectiveness of each component and perform extensive comparative experiments on four public benchmark datasets.

\end{abstract}

\begin{highlights}

\item We propose a dual-path feature extraction that integrates ResNet and DenseNet architectures to effectively extract complementary features from heterogeneous sensor modalities while achieving superior cross-modal semantic alignment.

\item  We adopt a confidence-driven gradient modulation mechanism that dynamically balances modality contributions during training by adaptively adjusting gradient intensities, preventing dominant modalities from suppressing weaker but complementary ones.

\item We built a multi-stage contrastive learning framework that performs progressive feature alignment from instance-level to semantic-level across multiple network layers, significantly enhancing cross-modal consistency and representation learning capability.

\end{highlights}

\begin{keywords}
 \sep Human Activity Recognition 
 \sep Cross-modal Alignment 
 \sep Gradient Modulation 
\end{keywords}
\maketitle

\section{Introduction}

Human Activity Recognition (HAR) refers to the task of automatically identifying and classifying human activities by analyzing data collected from sensors. This technology serves as a critical foundation for intelligent systems in terms of environmental perception and decision-making support.With the rapid advancement of wearable devices and the Internet of Things (IoT) technologies \cite{huang2024survey}, HAR has gained increasing attention in fields such as smart health monitoring \cite{chen2021apneadetector} and rehabilitation medicine \cite{zhang2020rehabilitation}.Based on the type of data input, HAR systems are generally categorized into vision-based and sensor-based approaches. Vision-based HAR systems capture human activity through video or image sequences obtained by cameras. These systems offer intuitive spatiotemporal features of body movements along with rich visual contextual information \cite{bayoudh2022survey}. However, their performance is often constrained by variations in actions, sensor characteristics, environmental conditions, data acquisition scope, and computational overhead \cite{yadav2021review}. Moreover, the reliance on fixed camera setups limits their applicability in mobile and privacy-sensitive environments. In contrast, sensor-based HAR systems utilize wearable or ambient-embedded sensors, such as accelerometers, gyroscopes, and magnetometers, to collect motion data in real time \cite{yin2024systematic}. These systems provide advantages in privacy protection, energy efficiency, and portability. Nonetheless, compared with vision-based systems, they typically operate with lower-dimensional data and lack the environmental context that can be valuable for behavior interpretation.

Earlier HAR research predominantly relied on conventional machine learning algorithms, including Support Vector Machines (SVM), Random Forests (RF), and K-Nearest Neighbors (KNN) \cite{bao2004activity}. These methods depend heavily on handcrafted feature design and manual selection, which limits their adaptability and generalization across diverse scenarios \cite{ravi2005activity}. To overcome these limitations, deep learning techniques have been introduced into the HAR domain. Models such as Convolutional Neural Networks (CNNs) \cite{zhang2022deep} and Long Short-Term Memory networks (LSTMs) \cite{uddin2021human} can automatically learn hierarchical feature representations from raw sensor data, offering superior capabilities in handling multimodal and large-scale data. As a result, deep learning has become the prevailing approach in contemporary HAR research \cite{wang2019deep}.

While unimodal HAR systems have achieved promising results in controlled environments, their performance often degrades in complex real-world scenarios due to an inability to fully capture the multidimensional nature of human activity \cite{yadav2021review}. To address this, multimodal HAR systems aim to fuse heterogeneous sensor data, leveraging complementary information across modalities to enhance both recognition accuracy and system robustness \cite{ehatisham2019robust}. Despite such advantages, multimodal HAR still faces two critical challenges:

\begin{enumerate}

\item \textbf{Modality contribution imbalance.} In multimodal HAR, different modalities contribute unequally to the recognition task \cite{das2023revisiting}. Dominant modalities tend to drive the overall optimization process, thereby suppressing the learning potential of weaker but complementary modalities. Furthermore, in some cases, newly added modalities offer limited improvement in accuracy, failing to fully exploit the value of multi-source data and reducing robustness and generalization in complex environments \cite{li2023boosting}.
\item \textbf{Difficulty in cross-modal feature alignment.} Signals from heterogeneous sensors differ significantly in terms of physical interpretation and statistical properties \cite{dufumier2024align}, making it extremely challenging to project them into a semantically consistent feature space. Existing approaches typically rely on static constraints or reconstruction-based objectives to enforce alignment. However, these strategies often fail to ensure semantic-level consistency, especially under real-world conditions where activity boundaries are ambiguous and environmental context is dynamic. Such misalignments not only hinder recognition accuracy but also compromise the system's generalization to unseen activity types, representing a core bottleneck in the scalability of multimodal HAR.
\end{enumerate}

The main contributions of this paper are summarized as follows:

\begin{itemize}
\item We propose a Dynamic Contrastive Dual-Path Network (DCDPN) that integrates dual-path feature extraction with multi-stage contrastive learning in a unified framework, enabling simultaneous optimization of cross-modal semantic alignment and modality contribution balance.
\item We adopt a Confidence-Driven Gradient Modulation mechanism, which adaptively adjusts the learning intensity of each modality branch based on prediction confidence, thereby reducing the interference of dominant modalities and effectively mitigating imbalance in multimodal learning.
\item We develop a Multi-Stage Contrastive Learning strategy to address semantic inconsistencies among heterogeneous sensor data in multimodal HAR systems, significantly improving the performance of cross-modal feature alignment.
\end{itemize}

The rest of the paper is organized as follows: Section 2 reviews the related work; Section 3 describes the proposed methodology in detail; Section 4 presents the experimental design and results analysis; Finally, Section 5 concludes the paper and discusses future research directions.

\section{Related Work}
\subsection{Modality Imbalance}

Modality imbalance has emerged as a major factor affecting model performance in multimodal learning. Wang et al.~\cite{wang2020makes} observed that on the Kinetics dataset, multimodal joint training could yield lower recognition accuracy than the best-performing unimodal models. This phenomenon was attributed to differences in learning speed across modalities, where faster-converging dominant modalities tend to steer the optimization process. To address this, Fujimori et al.~\cite{fujimori2020modality} proposed a Modality-Specific Learning Rate Control mechanism that dynamically adjusts learning rates based on the convergence speed of each modality's loss, effectively alleviating gradient domination.From the perspective of loss function design, Fan et al.~\cite{Fan_2023_CVPR} introduced the Prototypical Cross-Entropy (PCE) loss and Prototypical Entropy Regularization (PER) to promote cooperative convergence among modalities. In terms of training strategies, Zhang et al.~\cite{Zhang_2024_CVPR} proposed the Modality-wise Learning Alternation (MLA) scheme, which transforms joint multimodal training into sequential unimodal learning to reduce inter-modality interference. However, these methods generally rely on static balancing strategies, making them less adaptable to the dynamic nature of modality contributions during training.With the growing adoption of Transformer-based architectures in HAR, Ma et al.~\cite{ma2019attnsense} noted that conventional attention mechanisms often fail to accurately estimate the importance of features from different modalities, resulting in overemphasis on strong modalities. To mitigate this, they proposed AttnSense, a hierarchical attention mechanism that assigns attention weights across temporal, channel, and modality dimensions. Nevertheless, this method does not explicitly tackle modality imbalance from the perspective of training optimization.

Building upon these efforts, Peng et al.~\cite{Peng_2022_CVPR} proposed the OGM-GE framework, which dynamically monitors the contribution of each modality to the global optimization objective and adaptively regulates gradients. Additionally, a dynamic Gaussian noise injection mechanism is employed to enhance modality balance at a deeper level.Although notable progress has been made, existing approaches still exhibit limitations. Many methods~\cite{chung2019sensor,alharbi2022comparing,koupai2022self} are unable to adapt to real-time fluctuations in modality contributions during training, and strategies based solely on loss or learning rate adjustment fail to directly intervene in the gradient propagation process. To this end, we propose a Confidence-driven Gradient Modulation strategy that adaptively adjusts the gradient strength of each modality branch based on its predicted confidence. Combined with a momentum-based mechanism to ensure training stability, this method enables more precise and dynamic control of modality balance.

\subsection{Modality Alignment}

Modality alignment aims to establish semantic correspondences across different data modalities by projecting heterogeneous signals into a shared representation space, where semantically similar samples are brought closer together. This alignment is essential for cross-modal integration and semantic-level fusion.Early modality alignment techniques were largely based on linear assumptions. Canonical Correlation Analysis (CCA)~\cite{hotelling1992relations} maximized the correlation between projected representations from different modalities, enabling basic alignment. However, such linear methods are inadequate for capturing complex nonlinear relationships. To overcome this, Andrew et al.~\cite{andrew2013deep} introduced Deep CCA, and Wang et al.~\cite{wang2015deep} proposed shared representation learning frameworks based on deep networks. While these approaches enhanced expressive power through nonlinear mappings, their objective functions primarily focused on data reconstruction rather than explicit semantic alignment between modalities.The incorporation of attention mechanisms further improved alignment performance. Xu et al.~\cite{xu2015show} developed a cross-modal attention model that achieved soft alignment by learning dynamic attention weights. Similarly, Ma et al.~\cite{ma2019attnsense} designed a hierarchical attention mechanism that performs adaptive alignment at multiple abstraction levels. However, these approaches still lack direct optimization for semantic consistency across modalities.

Hjelm et al.~\cite{hjelm2018learning} proposed maximizing mutual information between modality-specific representations to capture statistical dependencies. However, estimating mutual information in high-dimensional spaces is notoriously difficult, leading to training instability and convergence issues. In contrast, contrastive learning has gained traction for avoiding these estimation challenges. Chen et al.~\cite{chen2020simple} extended contrastive learning to the multimodal domain by minimizing the distance between positive pairs and maximizing that of negative pairs, thus aligning semantic relationships across modalities. Radford et al.~\cite{radford2021learning} further demonstrated the power of large-scale contrastive learning through CLIP, achieving robust cross-modal alignment.Despite these advances, existing methods face two key limitations: (1) most models adopt a single network pathway to process all sensor modalities, neglecting the inherent differences among them; (2) alignment is typically performed only at the final feature layer, lacking semantic consistency across multiple levels of abstraction. To address these gaps, we propose a multi-stage contrastive learning framework based on dual-path feature extraction, which integrates a specialized network architecture and multi-level alignment strategies to significantly enhance modality alignment in HAR.

% Bibliography section (add your .bib file reference here)
% \bibliographystyle{plain}
% \bibliography{references}

\section{Methodology}
\begin{figure*}
    \centering
    \includegraphics[width=1\linewidth]{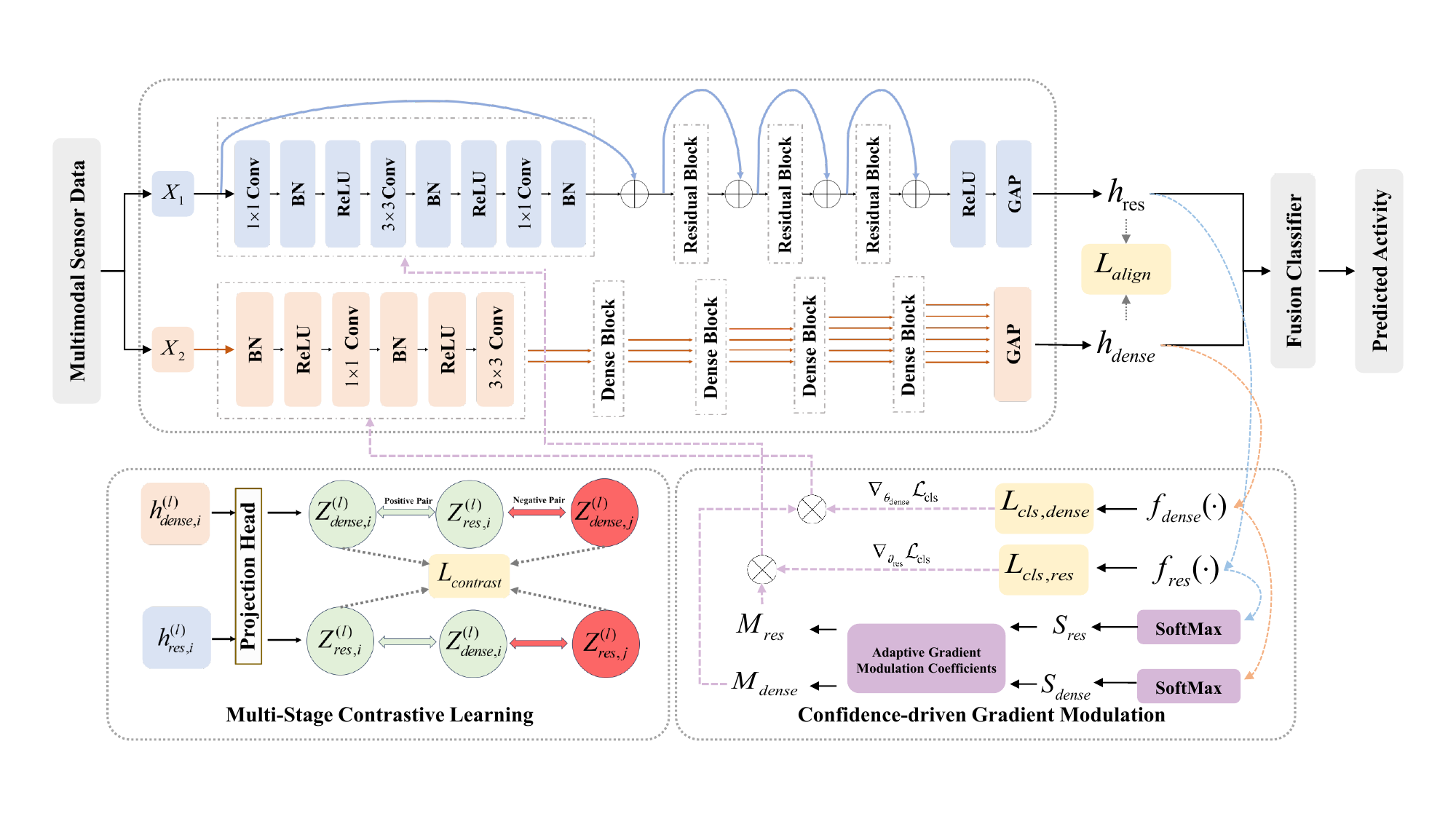}
    \caption{The total process of task. }
   \label{fig:flowchart}
\end{figure*}

This section introduces the proposed HAR framework. As shown in Figure \ref{fig:flowchart}.
To systematically address two fundamental challenges in multimodal HAR—cross-modal feature alignment difficulties and imbalanced modality contributions—we propose a novel end-to-end deep learning framework termed Dynamic Contrastive Dual-Path Network for Human Activity Recognition (DCDP-HAR). The framework comprises three interconnected components that collectively address these challenges:\par
\textbf{Dual-Path Feature Extraction}: This backbone component employs two parallel, architecturally complementary pathways that decouple and efficiently process multimodal sensor data exhibiting distinct physical properties.\par

\textbf{Multi-Stage Contrastive Learning}: This component performs feature alignment at multiple network levels. By maximizing the consistency between cross-modal positive pairs, it achieves progressive alignment from shallow, local features to deep, semantic features.\par

\textbf{Confidence-driven Gradient Modulation }: During backpropagation, this mechanism dynamically monitors and adjusts the learning intensity of each modality-specific branch to mitigate modal competition, thereby ensuring a balanced and efficient training dynamic.\par
These three components operate synergistically, forming a comprehensive solution that spans data processing, feature alignment, and training optimization. The overarching goal is to significantly enhance the accuracy and robustness of HAR systems in complex, real-world scenarios.

\subsection{Part A: Dual-Path Feature Extraction}
In multimodal HAR systems, signals acquired from heterogeneous sensor modalities exhibit substantial disparities in both physical interpretations and statistical characteristics. Conventional single-path architectures process all sensor channels uniformly, consequently neglecting this inherent heterogeneity and constraining feature extraction effectiveness. To address the heterogeneity of multimodal sensor data, we first formalize the input representation. \\
\indent Given an input tensor $\mathbf{X} \in \mathbb{R}^{N \times T \times F}$,where $N$ represents the batch size, $T$ denotes the temporal dimension, and $F$ indicates the feature dimension. We first partition
it into two complementary subsets:
$\boldsymbol{X}_1 \in \mathbb{R}^{N \times T \times F_1}$ and $\boldsymbol{X}_2 \in
\mathbb{R}^{N \times T \times F_2}$.
These subsets are determined by mutually exclusive feature index sets $\mathcal{I}_1$
and $\mathcal{I}_2$,
such that $\mathcal{I}_1 \cup \mathcal{I}_2 = \{1, \ldots, F\}$ and $\mathcal{I}_1
\cap \mathcal{I}_2 = \emptyset$, where $F_1 = |\mathcal{I}_1|$, $F_2 = |\mathcal{I}_2|$.
Recognizing that single-path networks struggle to balance feature reuse with new feature exploration, we propose a synergistic dual-path network architecture. This architecture leverages two complementary paths, ResNet and DenseNet, to achieve efficient feature learning.
The ResNet path-based, based on residual connections with identity mapping, defines its forward propagation for the k-th residual block as:
$$\mathbf{h}_{\text{res}}^{(k)} = \mathbf{h}_{\text{res}}^{(k-1)} + \mathcal{F}_k\left(\mathbf{h}_{\text{res}}^{(k-1)}; \theta_k^{\text{res}}\right)$$
where$\mathcal{F}_k(\cdot)$ denotes the $k$ -th residual mapping function, comprising batch normalization, ReLU activation, and convolutional operations. This design facilitates stable feature propagation and reuse, making it particularly suitable for processing core motion features with strong periodicity and stability, such as accelerometer signals.
The DenseNet-based pathway employs a dense cross-layer connection pattern, where each layer receives the feature maps of all preceding layers as input. The output of the k-th dense block is:
$$h_{\text{dense}}^{(k)} = \mathcal{H}_k\left(\left[h_{\text{dense}}^{(0)}, h_{\text{dense}}^{(1)}, \ldots, h_{\text{dense}}^{(k-1)}\right]; \theta_k^{\text{dense}}\right)$$
where $[\cdot]$represents concatenation along the feature dimension, and $\mathcal{H}_k(\cdot)$ is a composite non-linear transformation. This design promotes the exploration of new features through combinatorial reuse, enabling the capture of more fine-grained or unstructured signals.
Finally, compact feature representations are obtained from the last layer of each path via Global Average Pooling (GAP):
$$h_{\text{res}} = \text{GAP}(h_{\text{res}}^{(L_{\text{res}})}) \in \mathbb{R}^{N \times d_{\text{res}}}$$

$$h_{\text{dense}} = \text{GAP}(h_{\text{dense}}^{(L_{\text{dense}})}) \in \mathbb{R}^{N \times d_{\text{dense}}}$$
where $L_{\text{res}}$and$L_{\text{dense}} $are the number of blocks in the respective paths, and $d_{\text{res}} $and $d_{\text{dense}} $are the corresponding output feature dimensions.

\subsection{Part B: Multi-Stage Contrastive Learning}
To enforce semantic consistency across cross-modal features at multiple abstraction levels, we introduce a multi-stage contrastive learning mechanism. For each stage $l \in {1, 2, ..., L}$, we extract intermediate features 
$h_{\text{res}}^{(l)}$ and $h_{\text{dense}}^{(l)}$ from the two paths and map them to a shared embedding space via learnable projection heads:
$$z_{\text{res}}^{(l)} = g_{\text{res}}^{(l)}(h_{\text{res}}^{(l)}) \in \mathbb{R}^{N \times d_{\text{proj}}}$$

$$z_{\text{dense}}^{(l)} = g_{\text{dense}}^{(l)}(h_{\text{dense}}^{(l)}) \in \mathbb{R}^{N \times d_{\text{proj}}}$$ \\
where $g_{\text{res}}^{(l)}(\cdot)$ and $g_{\text{dense}}^{(l)}(\cdot)$ are non-linear projection functions that map features to a common space of dimension  $d_{\text{proj}}$.

During training, for a given activity sample within a batch, its projected features from the two paths,$(z_{\text{res}, i}^{(l)}, z_{\text{dense}, i}^{(l)})$ , constitute a positive pair. Features from different activity samples,$(z_{\text{res}, i}^{(l)}, z_{\text{dense}, j}^{(l)})$, where $i \neq j$, form negative pairs. We compute the cosine similarity matrix between the projected features:
$$\mathcal{S}_{i,j}^{(l)} = \frac{z_{\text{res}, i}^{(l) \top} z_{\text{dense}, j}^{(l)}}{\| z_{\text{res}, i}^{(l)} \|_2 \| z_{\text{dense}, j}^{(l)} \|_2}$$
Based on this similarity matrix, the bidirectional contrastive loss at stage $l$ is defined as:
$$\mathcal{L}_{\text{contrast}}^{(l)} = -\frac{1}{2N} \sum_{i=1}^N \left[ \log \frac{\exp(\mathcal{S}_{i,i}^{(l)} / \tau)}{\sum_{j=1}^N \exp(\mathcal{S}_{i,j}^{(l)} / \tau)} + \log \frac{\exp(\mathcal{S}_{i,i}^{(l)} / \tau)}{\sum_{j=1}^N \exp(\mathcal{S}_{j,i}^{(l)} / \tau)} \right]$$
where$\tau$is a temperature hyperparameter controlling the concentration of the distribution. The total multi-stage contrastive loss is the average of the losses over all stages:
$$\mathcal{L}_{\text{contrast}} = \frac{1}{L} \sum_{l=1}^L \mathcal{L}_{\text{contrast}}^{(l)}$$
Furthermore, to enforce consistency in the final feature representations, we employ a Mean Squared Error (MSE) loss to align the L2-normalized feature vectors from the two branches:
$$\mathcal{L}_{\text{align}} = \text{MSE} \left( \frac{h_{\text{res}}}{\| h_{\text{res}} \|_2}, \frac{h_{\text{dense}}}{\| h_{\text{dense}} \|_2} \right)$$

\subsection{Part C: Confidence-driven Gradient Modulation}
Although the dual-path network can effectively extract heterogeneous features from different sensor modalities via its ResNet and DenseNet sub-branches, the training process is often hampered by modality bias and gradient oscillation. Dominant modalities tend to monopolize the learning process, preventing the system from effectively leveraging complementary information from weaker modalities. To address this modality imbalance during training, we propose Confidence-driven Gradient Modulation (CGM), which achieves balanced multimodal learning by dynamically adjusting the learning intensity of each branch.
For a sample $(x_i, y_i)$in a batch, we first compute the prediction confidences for the true class$y_i$ from the classifiers of both branches, $f_{\text{res}}(\cdot)$ and $f_{\text{dense}}(\cdot)$. The total confidences for the entire batch are then accumulated:

$$S_{\text{res}} = \sum_{i=1}^N \text{Softmax}(f_{\text{res}}(x_{i,1}))[y_i]$$
$$S_{\text{dense}} = \sum_{i=1}^N \text{Softmax}(f_{\text{dense}}(x_{i,2}))[y_i]$$
where $x_{i,1}$ and $x_{i,2}$ are the inputs to the ResNet-based and DenseNet-based pathways, respectively.
Based on these confidence sums, we define modality contribution ratios to quantify the relative dominance between modalities:

$$R_{\text{res}} = \frac{S_{\text{res}}}{S_{\text{dense}} + \epsilon}, \quad R_{\text{dense}} = \frac{S_{\text{dense}}}{S_{\text{res}} + \epsilon}$$
where $\epsilon$ is a small positive constant to prevent division by zero. Subsequently, we design adaptive gradient modulation coefficients,$M_{\text{res}}$ and $M_{\text{dense}}$, to suppress over-contributing modalities: 
$$M_{\text{res}} = 
\begin{cases} 
1 - \tanh(\alpha \cdot \text{ReLU}(R_{\text{res}} - 1)), & \text{if } R_{\text{res}} > 1 \\
1, & \text{otherwise}
\end{cases}$$

$$M_{\text{dense}} = 
\begin{cases} 
1 - \tanh(\alpha \cdot \text{ReLU}(R_{\text{dense}} - 1)), & \text{if } R_{\text{dense}} > 1 \\
1, & \text{otherwise}
\end{cases}$$
where $\alpha$is a hyperparameter controlling the modulation strength. During backpropagation, these coefficients directly scale the gradients of the respective branch classifiers:

$$\tilde{\nabla}_{\theta_{\text{res}}} \mathcal{L}_{\text{cls}} = M_{\text{res}} \cdot \nabla_{\theta_{\text{res}}} \mathcal{L}_{\text{cls}}$$

$$\tilde{\nabla}_{\theta_{\text{dense}}} \mathcal{L}_{\text{cls}} = M_{\text{dense}} \cdot \nabla_{\theta_{\text{dense}}} \mathcal{L}_{\text{cls}}$$
This mechanism dynamically suppresses the learning rate of the dominant modality, affording weaker modalities a greater opportunity to learn, thus achieving a balanced training objective.
The introduction of CGM may lead to oscillations in the gradient direction. To enhance training stability, we employ a momentum-based gradient accumulation strategy to smooth the optimization path. By integrating historical gradient information, momentum effectively dampens drastic changes in the update direction, ensuring stable convergence for each branch under dynamic adjustment. A general momentum update rule is:

$$m_t = \beta m_{t-1} + (1 - \beta) \tilde{g}_t$$
$$\theta_{t+1} = \theta_t - \eta_t m_t$$
where $m_t$ is the momentum at timestep $t$,$\tilde{g}_t$is the CGM-modulated gradient, $\beta$is the momentum coefficient, and $\eta_t$ is the learning rate. This stabilization strategy, combined with CGM, ensures that the model maintains efficient and robust convergence while balancing the learning intensities of different modalities.

\subsubsection{Total Training Objective}
The final training objective of our model is a multi-task loss function that combines three classification losses with the alignment and contrastive learning losses. The total loss$\mathcal{L}_\text{total}$is defined as:
$$\mathcal{L}_\text{total} = \left( \mathcal{L}_\text{cls,res} + \mathcal{L}_\text{cls,dense} + \mathcal{L}_\text{cls,fusion} \right) + \lambda_\text{align} \left( \mathcal{L}_\text{contrast} + \mathcal{L}_\text{align} \right)$$
The hyperparameter $\lambda_\text{align}$ balances the classification objective with the feature alignment tasks. ${L}_\text{cls,res}$ and ${L}_\text{cls,dense}$represent the classification losses for the ResNet branch and DenseNet branch, respectively. This integrated design achieves cross-modal semantic alignment through multi-stage contrastive learning while ensuring a balanced training process via the CGM mechanism, leading to efficient multimodal feature learning and superior classification performance.

\section{Experimental Design}
All experiments were performed on a high-performance server featuring seven NVIDIA Tesla V100-SXM3 GPUs (32GB memory each), an AMD EPYC 7H12 64-core processor, and 1024GB RAM. The software environment was based on Python 3.12.3 and PyTorch 2.6.0.

\subsection{Data Sets Used in the Experiment}
In order to ensure an objective evaluation of our methodology, several pertinent aspects of these used datasets have been outlined below. \\ 
\indent OPPORTUNITY\cite{chavarriaga2013opportunity}captures diverse daily activities performed by 12 participants, including walking, standing, sitting, and other routine behaviors. The comprehensive sensor infrastructure encompasses 15 networked sensor systems comprising 72 individual sensors across 10 distinct modalities, including body-worn inertial sensors, object-integrated sensors, and ambient environmental sensors.
PAMAP2\cite{reiss2012introducing}contains recordings of 9 participants executing 18 different physical activities encompassing locomotion, household tasks, and sports-related movements. The multimodal sensor configuration consists of three Inertial Measurement Units (IMUs) strategically positioned on participants' dominant hand, chest, and ankle. Each IMU integrates a tri-axial accelerometer, gyroscope, and magnetometer.
WISDM\cite{kwapisz2011activity}focuses on smartphone-based activity recognition using accelerometer data, comprising recordings from 36 participants performing six fundamental daily activities: walking, jogging, ascending stairs, descending stairs, sitting, and standing. Data collection utilized smartphone accelerometers operating at 20 Hz sampling frequency.
UCI-HAR\cite{anguita2013public}contains sensor recordings from 30 participants performing six activities of daily living using waist-mounted smartphones. Each device captured tri-axial linear acceleration and tri-axial angular velocity signals at 50 Hz frequency through embedded accelerometers and gyroscopes.

\subsection{Evaluation Metrics}
To quantitatively evaluate the performance of the model, we adopted the following evaluation metrics widely used in the HAR :\\
\indent Accuracy, the core metric of interest, measures the proportion of correct classifications made by the model across all samples, providing a quick overview of overall performance. F1-weighted offers a synthesis of performance for multiclass classification. The combined use of these metrics allows for a thorough understanding of the model's strengths and limitations in the task of HAR. 
$$
\text{Accuracy} = \frac{\mathrm{TP} + \mathrm{TN}}{\mathrm{TP} + \mathrm{FN} + \mathrm{FP} + \mathrm{TN}}
$$

$$
\text{Precision} = \frac{\mathrm{TP}}{\mathrm{TP} + \mathrm{FP}}
$$

$$
\text{Recall} = \frac{\mathrm{TP}}{\mathrm{TP} + \mathrm{FN}}
$$

$$
\text{F1\text{-}macro} = \frac{2 \times (\text{Precision} \times \text{Recall})}{\text{Precision} + \text{Recall}}
$$

$$
\text{F1\text{-}weighted} = \sum_{i} \frac{2 \times \omega_i \times (\text{Precision}_i \times \text{Recall}_i)}{\text{Precision}_i + \text{Recall}_i}
$$

$$
\text{F1-score} = \frac{2 \times \mathrm{TP}}{2 \times \mathrm{TP} + \mathrm{FP} + \mathrm{FN}}
$$

Where TP and TN  represent the number of true positives and true negatives, respectively, while FN and FP are the numbers of false negatives and false positives. Precision represents the average precision across all labels, and Recall represents the average recall rate across all labels.  $\omega_{i}$ is the proportion of the $i$ class samples. 

\section{Results and Discussion}
\subsection{Hyperparameter Tuning and Model Optimization}
To optimize DCDP-HAR performance, we conducted systematic hyperparameter tuning experiments across multiple stages. This optimization was conducted on the OPPO dataset.
\begin{table*}
\centering
\renewcommand\arraystretch{1.3} % 调整行高
\tabcolsep=0.5cm % 调整列间距

\caption{Parameter configurations for different experimental stages. This is a staged parameter optimization process. In each stage, other relevant parameters are set to the optimal values determined in the previous stages.}
\label{tab:param_config}

\begin{tabular}{c | c | c c} 
\toprule
\textbf{Stage} & \textbf{Varied Parameter} & \textbf{Group ID} & \textbf{Value} \\
\cmidrule(r){1-4}

\multirow{5}{*}{\textbf{\makecell{Optimizer \\ Tuning}}}
& \multirow{5}{*}{Optimizer + Learning Rate} 
& A1 & AdamW, lr=0.001 \\
& & A2 & AdamW, lr=0.0005 \\
& & A3 & Adam, lr=0.001 \\
& & A4 & SGD, lr=0.001 \\
& & A5 & SGD, lr=0.0005 \\
\midrule 

\multirow{4}{*}{\textbf{\makecell{Architecture \\ Tuning}}}
& \multirow{4}{*}{\makecell{ResNet \& DenseNet \\ Layers}} 
& B1 & ResNet: [1,1,1,1], DenseNet: [2,2,2,2] \\
& & B2 & ResNet: [2,2,2,2], DenseNet: [3,3,4,3] \\
& & B3 & ResNet: [3,4,6,3], DenseNet: [4,4,6,4] \\
& & B4 & ResNet: [3,4,23,3], DenseNet: [6,8,12,6] \\
\midrule 

\multirow{5}{*}{\textbf{\makecell{Weight \& Temp \\ Tuning}}}
& \multirow{5}{*}{Weight \& Temperature}
& C1 & Weight=0.1, Temp=0.3 \\
& & C2 & Weight=0.3, Temp=0.5 \\
& & C3 & Weight=0.5, Temp=0.7 \\
& & C4 & Weight=0.7, Temp=0.5 \\
& & C5 & Weight=1.0, Temp=0.3 \\
\midrule

\multirow{5}{*}{\textbf{\makecell{CGM Alpha \\ Tuning}}}
& \multirow{5}{*}{CGM Alpha} 
& D1 & 0.1 \\
& & D2 & 0.3 \\
& & D3 & 0.5 \\
& & D4 & 0.7 \\
& & D5 & 0.9 \\

\bottomrule
\end{tabular}
\end{table*}

\subsection{Core Parameter Tuning}
For preliminary hyperparameter optimization, we evaluated three commonly used optimizers: AdamW, Adam, and Stochastic Gradient Descent (SGD) with momentum.For initial evaluation, we employed a dual-path network architecture of relatively low complexity, with ResNet branches configured as [1,2,2,1] and DenseNet branches as [2,2,3,2].\par
Configuration A1 demonstrated superior performance across all evaluated settings, achieving the highest mean test accuracy of  88.12\% (±0.08) and the optimal F1-Macro and F1-Weighted scores of 0.880 and 0.882, respectively. These results indicate that AdamW's adaptive learning rate mechanism proves more effective for our optimization task, delivering superior parameter updates across model components while ensuring faster convergence and enhanced robustness. In contrast, all SGD configurations and other Adam configurations exhibited inferior performance across all evaluation metrics.\par
We subsequently evaluated four distinct architectural configurations to determine the optimal depth and width for the ResNet and DenseNet components within our model. This evaluation aimed to establish a backbone network with sufficient representational capacity to support subsequent cross-modal feature alignment and modal balance adjustments.
The configuration B3 achieved the highest average performance across both test accuracy and F1-weighted metrics, attaining 87.67\% and 0.878, respectively. This configuration effectively increased model capacity to capture the inherent complexity of the dataset while maintaining robust performance compared to the configuration B4 with the largest parameter count, which demonstrated superior handling of overfitting risks and diminishing returns.
Although the F1-macro score for this optimal configuration was marginally lower than that achieved by the configuration B2, the latter configuration exhibited a notable decline in overall accuracy performance. These results indicate that configuration B3 provides an effective balance between model complexity and generalization capability, making it the preferred architectural foundation for our subsequent analyses.
\begin{figure}
    \centering
    \includegraphics[width=1\linewidth]{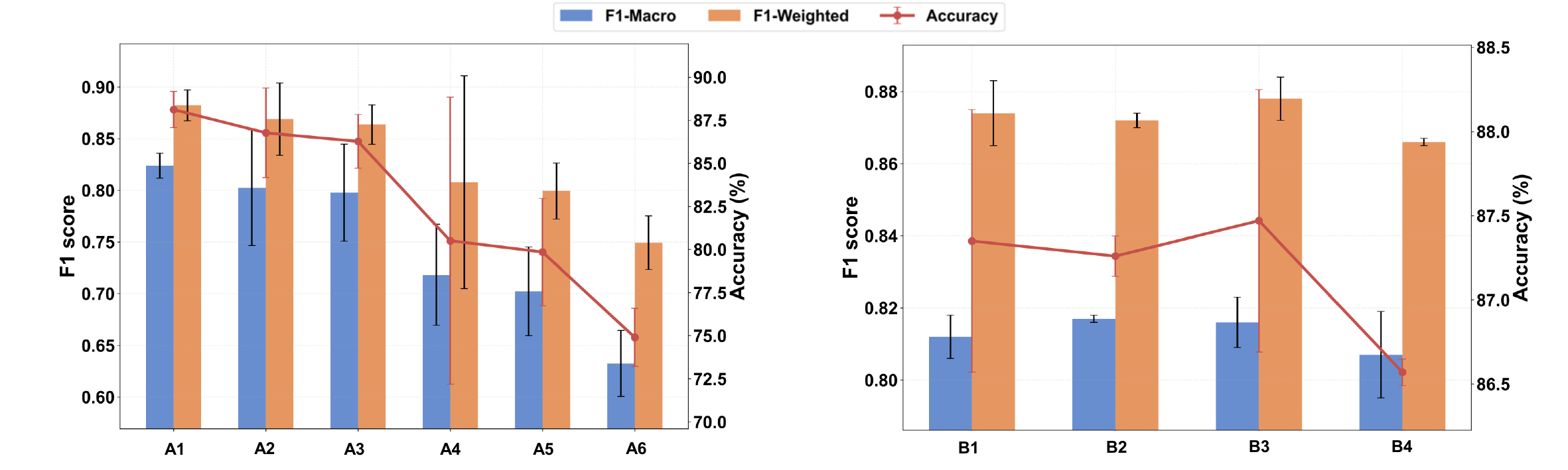}
    \caption{Optimizer and Network Architecture Tuning (Group A \& Group B) on Performance Metrics of the OPPORTUNITY Dataset}
    \label{fig:Core Parameter Tuning1}
\end{figure}

Building upon the selected optimal architecture, we conducted experiments on five distinct combinations of weight and temperature parameters, with specific configurations presented on Table\ref{tab:param_config}. The experimental results demonstrate that optimal model performance was achieved when the contrastive learning weight was set to 0.7 and the temperature parameter to 0.5, yielding an accuracy of 87.80\%. \par
Compared to the best results obtained using only the baseline network structure, this configuration resulted in an accuracy improvement of 0.13\%. These findings substantiate that contrastive learning enables more effective fusion and understanding of features across different modalities, thereby enhancing performance in downstream classification tasks. \par
The temperature parameter (T = 0.5) controls the sharpness of the similarity distribution, while the weight of 0.7 establishes an optimal balance between the primary classification task and the auxiliary feature alignment task. This parameter combination provides the optimal equilibrium for feature alignment tasks in our study, promoting more consistent cross-modal representations and facilitating improved model performance through enhanced inter-modal feature coherence. \par
These results confirm that the incorporation of contrastive learning with appropriately tuned parameters significantly contributes to the model's ability to learn robust cross-modal representations, ultimately leading to superior classification performance.

\begin{figure}
    \centering
    \includegraphics[width=1\linewidth]{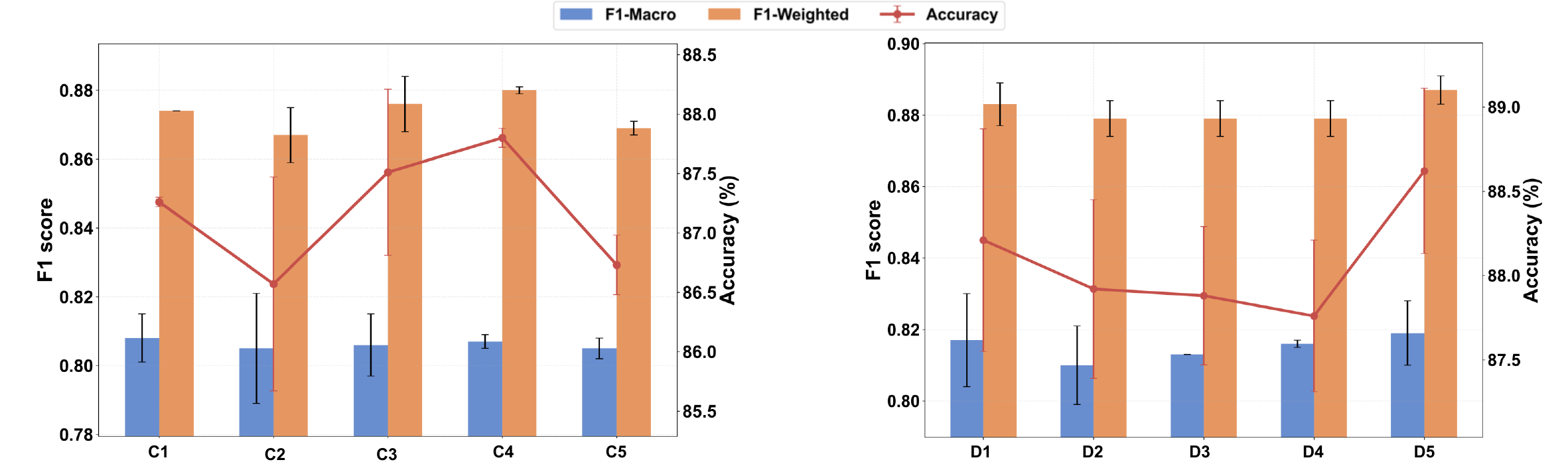}
    \caption{Contrastive Learning and CGM Parameter Tuning (Group C \& Group D) on Performance Metrics of the OPPORTUNITY Dataset}
    \label{fig:Core Parameter Tuning2}
\end{figure}

We subsequently investigated a comprehensive range of CGM Alpha values from 0.1 to 0.9, with experimental results presented in Figure \ref{fig:Core Parameter Tuning2}. The CGM mechanism operates by modifying gradients during the backpropagation process, thereby reducing the gradient influence of dominant modalities and providing enhanced optimization opportunities for modalities that demonstrate slower learning dynamics or exhibit less prominent contributions to the overall learning objective. \par
Figure \ref{fig:Core Parameter Tuning2} clearly demonstrates that model performance improved systematically as $\alpha$ values increased across the tested range. When Alpha was configured to 0.9, the model achieved the highest performance recorded in this investigation, with a mean test accuracy of 88.62\% (±0.49) and an F1-Weighted score of 0.887 (±0.004). This configuration yielded a substantial accuracy improvement of 0.82\% compared to the baseline model utilizing contrastive learning alone. \par
The optimal performance achieved with a high $\alpha$  value (0.9) indicates that stronger orthogonalization adjustments applied to gradients effectively facilitate balanced learning across all modalities. \par
Following the systematic staged parameter exploration and optimization process described above, we established the definitive optimal configuration for DCDP-HAR. This finalized configuration was subsequently employed for all ablation studies and comparative performance evaluations, ensuring consistent experimental conditions and reliable benchmark comparisons throughout the remainder of our investigation.\par
\subsection{Ablation Study}
We performed comprehensive ablation studies on the DCDP-HAR framework to assess individual component contributions. In these investigations, we systematically removed the dual-branch module, gradient modulation module, and feature alignment module, respectively, to evaluate their individual contributions to overall model performance. The experimental evaluations were performed on both the PAMP2 and OPPO datasets, with detailed results presented in Table \ref{tab:ablation_study}

\begin{table*}[t!]
    \centering
    \caption{Ablation study }
    \label{tab:ablation_study}
    \renewcommand\arraystretch{1.15}
    \begin{tabularx}{\linewidth}{lcccc *{4}{>{\centering\arraybackslash}X}}
        \toprule
        \multirow{2}{*}{\textbf{Method}} & \multicolumn{4}{c}{\textbf{Components}} & \multicolumn{4}{c}{\textbf{Performance Metrics}} \\
        \cmidrule(r){2-5} \cmidrule(r){6-9}
         & DPFE & CL & CGM & DA & ACC (\%) & Precision & Recall & F1-Score \\
        \midrule
        
        Baseline (ResNet) & \ding{55} & \ding{55} & \ding{55} & \ding{55} & 85.52 & 0.8713 & 0.8552 & 0.8536 \\
        \midrule
        + DPFE & \checkmark & \ding{55} & \ding{55} & \ding{55} & 91.71 & 0.9240 & 0.9171 & 0.9134 \\
        + DPFE + CL & \checkmark & \checkmark & \ding{55} & \ding{55} & 91.24 & 0.9138 & 0.9124 & 0.9104 \\
        + DPFE + CGM & \checkmark & \ding{55} & \checkmark & \ding{55} & 91.33 & 0.9176 & 0.9133 & 0.9108 \\
        + DP + DA & \checkmark & \ding{55} & \ding{55} & \checkmark & 91.62 & 0.9183 & 0.9162 & 0.9131 \\
        + DPFE + CL + CGM & \checkmark & \checkmark & \checkmark & \ding{55} & 92.00 & 0.9186 & 0.9200 & 0.9164 \\
        + DPFE + CGM + DA & \checkmark & \ding{55} & \checkmark & \checkmark & 92.86 & 0.9341 & 0.9286 & 0.9274 \\
        + DPFE + CL + DA & \checkmark & \checkmark & \ding{55} & \checkmark & 90.57 & 0.9110 & 0.9057 & 0.8994 \\
        \midrule
        \textbf{DCDP-HAR} & \checkmark & \checkmark & \checkmark & \checkmark & \textbf{93.90} & \textbf{0.9430} & \textbf{0.9390} & \textbf{0.9380} \\
        \bottomrule
        \multicolumn{9}{l}{\parbox[t]{\linewidth}{\footnotesize{Note: The ResNet baseline utilizes a single ResNet-based pathway for feature extraction, CL denotes Multi-Stage Contrastive Learning, DA denotes Data Augmentation.}}}
    \end{tabularx}
\end{table*}
Results demonstrate that the dual-path architecture substantially improves all performance metrics, with accuracy improving from 85.52\% to 91.71\% on PAMP2. This outcome provides strong evidence that the ResNet and DenseNet branches can learn complementary feature representations. Removing any single module leads to performance degradation, resulting in approximately 3\% accuracy reduction on the PAMAP2 dataset. This indicates that the DCDP-HAR model, through the synergistic effects of gradient modulation, feature alignment, and momentum updates, can more accurately predict activity types that other models struggle to identify.\par
After removing the contrastive learning module, accuracy decreased by 1.04\% on the PAMAP2 dataset. This confirms that our contrastive learning mechanism successfully constructs a more consistent and discriminative shared feature space. Following the removal of the CGM module, accuracy decreased from 93.90\% to 90.57\%. This clearly demonstrates that CGM effectively balances the learning process across modalities, ensuring the model can collaboratively learn from all available data sources. This not only enhances the model's final performance but also strengthens its robustness when facing different sensor combinations.\par
Specifically, removing the gradient modulation module weakens the model's capability to handle class imbalance problems, while removing the alignment module disrupts the model's ability to generate more discriminative joint representations. This series of performance gains strongly confirms that our model can effectively extract deep discriminative features and successfully alleviate gradient conflicts in multi-task learning, thereby validating the accuracy of our core design assumptions.

\begin{figure}
    \centering
    \includegraphics[width=1\linewidth]{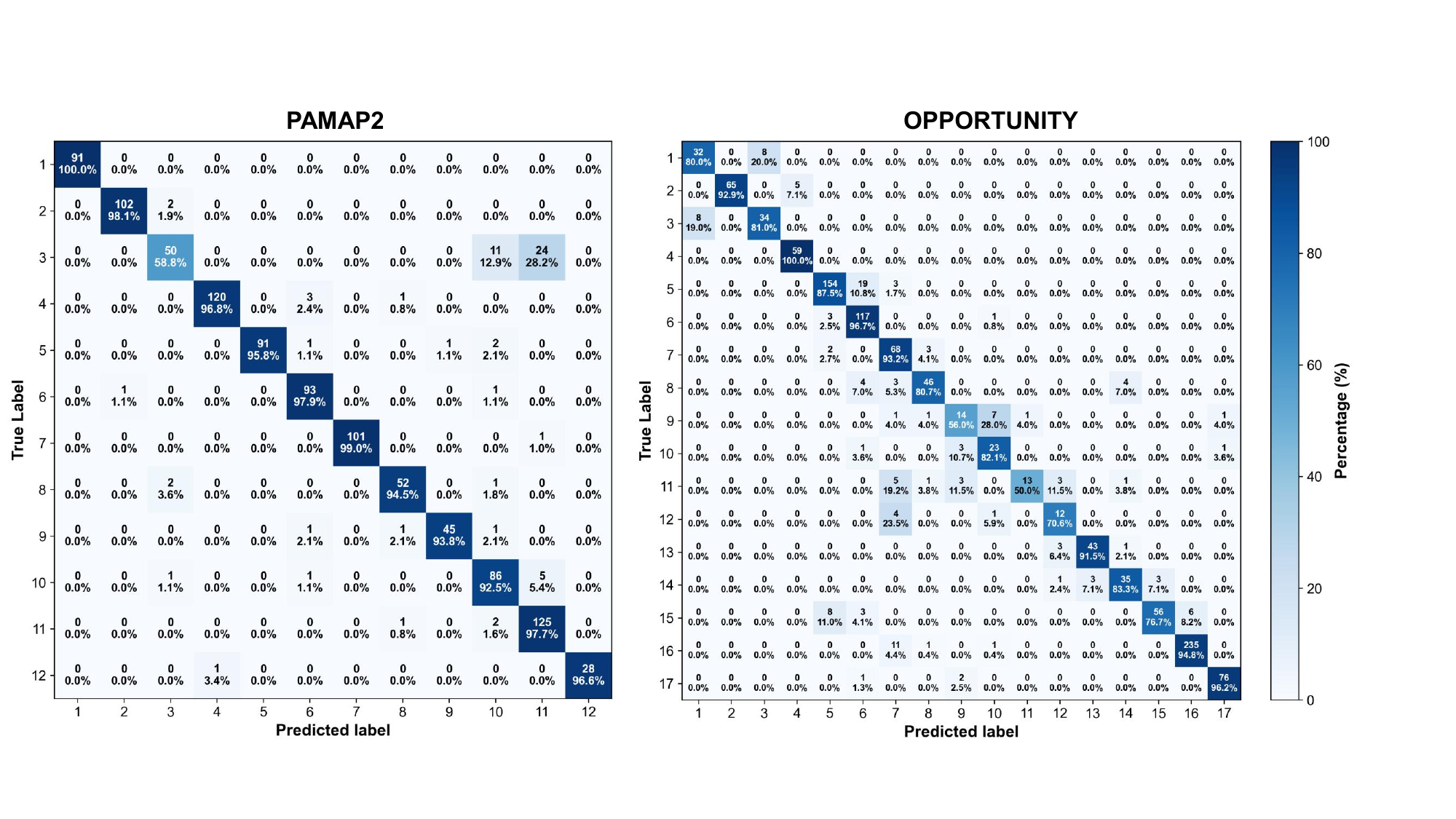}
    \caption{Confusion matrices on the PAMAP2 and OPPORTUNITY}
    \label{fig:ConfusionMatrix}
\end{figure}

The confusion matrices of DCDP-HAR on the OPPORTUNITY and PAMAP2 datasets further intuitively demonstrate its superior fine-grained recognition capability as shown in Figure \ref{fig:ConfusionMatrix}. DCDP-HAR achieves higher accuracy across more categories and effectively reduces problems caused by class confusion. From the confusion matrix of the OPPORTUNITY dataset, it can be observed that the model maintains extremely high recognition accuracy across the vast majority of categories, with values on the diagonal generally exceeding 90\%. Similarly, in the confusion matrix of the PAMAP2 dataset, the model also performs excellently. High confusion exists between categories "2", "3", "10", and "11", while DCDP-HAR optimizes this to a certain extent. These results strongly demonstrate that the model can effectively capture and distinguish subtle spatial-temporal features of different activities. Even in complex scenarios with numerous categories and high similarity between actions, it can significantly alleviate class confusion problems. Overall, through sophisticated architectural design and optimization strategies, our model exhibits powerful comprehensive performance when handling extremely challenging human activity recognition tasks.

\subsection{Comparison of related work}
To evaluate the performance of DCDP-HAR in HAR tasks, we compare it with a series of existing methods, covering classical deep learning frameworks and recent advanced models. All experiments are conducted on four widely used public datasets to ensure fair and comprehensive evaluation. The test results are shown in Table \ref{tab:performance_comparison}. \\
We first compare DCDP-HAR with foundational deep learning architectures that serve as cornerstones of HAR research. As presented in Table 3, traditional methods including CNN\cite{zeng2014convolutional}, LSTM\cite{xia2020lstm}, and their hybrid architectures\cite{xia2020lstm}\cite{dua2021multi} exhibit significant performance limitations across all test datasets. On the most challenging Opportunity dataset, which features complex multimodal sensor configurations and severe class imbalance issues, DCDP-HAR achieves 88.34\% accuracy, improving by 6.19\% and 6.69\% compared to CNN and LSTM respectively. This significant improvement demonstrates the critical importance of addressing modal imbalance in complex HAR scenarios, where traditional methods fail to effectively utilize complementary information between different sensor modalities. More importantly, the internal cross-modal contrastive learning mechanism begins modal alignment in the early stages of feature learning, avoiding information loss that may occur in late-stage fusion. Similar performance improvements are validated on WISDM, UCI-HAR, and OPPO datasets.\\
\begin{table*}
\centering
\renewcommand\arraystretch{1.2}
\tabcolsep=0.3cm
\caption{Performance comparison of different models on various datasets}
\label{tab:performance_comparison}
\begin{tabular}{c|cccc}     
\toprule
\textbf{Model} & \textbf{Opportunity} & \textbf{WISDM} & \textbf{UCI-HAR} & \textbf{PAMAP2} \\ 
\cmidrule(r){1-1}\cmidrule(r){2-2}\cmidrule(r){3-3}\cmidrule(r){4-4}\cmidrule(r){5-5}
CNN & 82.15 & 93.31 & 92.39 & 90.96 \\
\cmidrule(r){1-1}\cmidrule(r){2-2}\cmidrule(r){3-3}\cmidrule(r){4-4}\cmidrule(r){5-5}
LSTM & 81.65 & 96.71 & 95.52 & 91.51 \\
\cmidrule(r){1-1}\cmidrule(r){2-2}\cmidrule(r){3-3}\cmidrule(r){4-4}\cmidrule(r){5-5}
LSTM-CNN & 77.64 & 95.90 & 97.01 & 91.43 \\
\cmidrule(r){1-1}\cmidrule(r){2-2}\cmidrule(r){3-3}\cmidrule(r){4-4}\cmidrule(r){5-5}
CNN-GRU & 79.85 & 94.95 & 95.11 & 91.58 \\
\cmidrule(r){1-1}\cmidrule(r){2-2}\cmidrule(r){3-3}\cmidrule(r){4-4}\cmidrule(r){5-5}
ELK & 87.96 & 98.05 & 97.25 & 95.53 \\
\cmidrule(r){1-1}\cmidrule(r){2-2}\cmidrule(r){3-3}\cmidrule(r){4-4}\cmidrule(r){5-5}
DanHAR & 86.16 & 94.35 & 91.44 & 92.18 \\
\cmidrule(r){1-1}\cmidrule(r){2-2}\cmidrule(r){3-3}\cmidrule(r){4-4}\cmidrule(r){5-5}
DCDP-HAR & \textbf{88.34} & \textbf{98.75} & \textbf{97.79} & \textbf{93.71} \\
\bottomrule
\end{tabular}
\end{table*}

Compared with DanHAR\cite{gao2021danhar}, which employs dual attention networks for multimodal sensor fusion, we achieve competitive results. DCDP-HAR outperforms DanHAR by 2.18\% on the Opportunity dataset and by 4.40\% on WISDM. This comparison highlights the advantages of our dual-path architecture combined with dynamic gradient modulation over pure attention mechanisms.\\

The ELK method\cite{gao2021danhar}, representing the latest advances in CNN architecture design for HAR, demonstrates strong performance across datasets. However, DCDP-HAR consistently outperforms ELK, particularly in the most challenging scenarios, achieving excellent results on Opportunity and WISDM. This comparison validates the effectiveness of our modal balancing strategy compared to pure architectural improvements.

\section{Conclusion}
This study presents DCDP-HAR, a comprehensive sensor-based Human Activity Recognition framework that effectively addresses the fundamental challenges of cross-modal feature alignment and modality imbalance in HAR systems. The proposed methodology integrates dual-path feature extraction based on ResNet and DenseNet architectures, multi-stage contrastive learning, and confidence-driven gradient modulation to achieve superior recognition performance. Our comparative experiments across multiple datasets demonstrate the effectiveness of DCDP-HAR. Comprehensive ablation studies confirm that each component contributes significantly to overall performance, with the complete framework exhibiting substantial improvements over baseline configurations.

\printcredits

\bibliographystyle{elsarticle-num}

\bibliography{cas-refs}

\end{document}